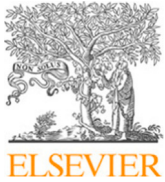
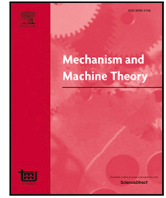

Research paper

# Direct kinematics, inverse kinematics, and motion planning of 1-DoF rational linkages

Daniel Huczala [a],[*], Andreas Mair [a], Tomas Postulka [b]

[a] *Unit of Geometry and Surveying, University of Innsbruck, Technikerstraße 13, Innsbruck, 6020, Austria*
[b] *Department of Robotics, VSB – Technical University of Ostrava, 17. listopadu 2172/15, Ostrava, 70800, Czech Republic*

ABSTRACT

This study presents a set of algorithms that deal with trajectory planning of rational single-loop mechanisms with one degree of freedom (DoF). Benefiting from a dual quaternion representation of a rational motion, a formula for direct (forward) kinematics, a numerical inverse kinematics algorithm, and the generation of a driving-joint trajectory are provided. A novel approach using the Gauss–Newton search for the one-parameter inverse kinematics problem is presented. Additionally, a method for performing smooth equidistant travel of the tool is provided by applying arc-length reparameterization. This general approach can be applied to one-DoF mechanisms with four to seven joints characterized by a rational motion, without any additional geometrical analysis. An experiment was performed to demonstrate the usage in a laboratory setup.

## 1. Introduction

Single-loop closed-chain spatial linkages belong to the group of mechanisms with *n*-number joint-link pairs that are consecutively connected in a loop. Often, they are referred to as *n*-bar linkages; the best-known example is probably the Bennett mechanism [1], which is the only moveable spatial 4-bar linkage. The reason for using linkages with one degree of freedom (DoF) is that they can perform, as custom mechanisms, complicated trajectories [2] with only a single motor. Therefore, similar to parallel robots, they can be more energy efficient than serial robots [3,4], which are now the standard in industry as manipulators with six DoF. However, the synthesis and design of single-loop mechanisms may be complicated. This paper presents the state-of-the-art approach to designing such linkages and novel algorithms for their control. We will focus on linkages with revolute (R) joints, as their mechanical design in terms of closed-loop structures is simpler and more common compared to prismatic or cylindrical joints, and importantly due to the recent study by Li et al. [5] they can be designed to avoid self-intersections, i.e., links and joints never collide during the motion.

The first challenge in designing 4R to 6R linkages is their over-constrained [6] nature. As a consequence, the Grübler–Kutzbach–Chebyshev formula for calculating the theoretical DoFs of the mechanism yields 0 for 6R linkages and −2 for 4R linkages. This means that such mechanisms should be rigid or not even able to be assembled. The over-constrained linkages move because of their specific geometric design constraints; however, the application of standard design and control algorithms may be tricky or sometimes even impossible for various reasons as, for instance, the Jacobian of such linkages is always singular. Nevertheless, fewer joints and links simplify the design which may be sufficient in basic pick-and-place manipulation. If one designs a mechanism that is characterized by a *rational* motion, it may benefit from existing algebraic methods that simplify the whole process.

Rational single-loop linkages are a fully classified [7] family of mechanisms and, by nature, they possess a continuous rational *parameterization* of the motion in polynomial form. Parameterization brings many advantages, such as the construction of motion polynomials from poses [1,2] or points [8] (task-based design), synthesis of kinematic structure using the rational motion factorization method [9], or search for collision-free realizations [5]. All of these methods are algorithmic, and their implementation and practical use in the design of linkage prototypes is presented in [10], the author's previous research. This paper is a continuation of work on *rational linkages* started in 2012 [11], and presents another step to bring the deep theoretical approach of modern linkage design closer to industrial applications.

As the authors are aware, there are no studies that follow up on the results of kinematic synthesis and design of single-loop spatial *n*-bar linkages – whether rational or non-rational – that provide a general algorithm for their kinematic control. There exist methods used to control *planar* 4-bars, however, these mechanisms are not over-constrained and standard methods may be applied. For example, Çakar and Tanyıldızı [12] use the approach via the Jacobian to determine velocities and accelerations, which is impossible in the case of spatial 4R linkages, as their Jacobian is always rank deficient due to the over-constrained property [13]. Yans [14] avoid using the Jacobian, but at the cost of focusing purely on the velocities of the driving joint. Additionally, the simplified dynamics model of the planar 4-bar suffers from redundant constraints in the case of a spatial 4-bar.

This paper briefly explains the direct kinematics of rational linkages, which is, for any part of the mechanism, straightforward – a rare and advantageous property for an arbitrary closed-loop kinematic structure. Additionally, it provides a solution for the velocity kinematics motion planning of rational 1-DoF linkages and related methods, most importantly, a novel algorithm to calculate the numerical inverse kinematics and a trajectory generation algorithm for a smooth end-effector path motion. The methodology shares similarities with the standard approach using the Jacobian; however, it completely avoids its calculation (as it would be rank deficient anyway), effectively utilizing the rational representation instead. The results are demonstrated in an experiment (see video in [15]). The algorithms are general and can be applied to any 1-DoF (mostly 4 to 7-bar) linkage whose motion is rational. For example, the Bennett linkages (all spatially movable 4-bars) are rational [16], and the motion curve can be recovered from the already existing design.

## 2. Mathematical background

This chapter briefly explains the theoretical framework behind the proposed methodology.

### 2.1. Dual quaternions

The group of proper spatial motions is called the special Euclidean group SE(3) and can be expressed in various ways; the most common interpretation is probably the representation as transformation matrices. In this paper, we use another geometrical representation that uses dual quaternions, also known as Study parameters. Dual quaternions $\mathbb{DH}$ are an extension of quaternions in the following way. Given quaternions $p_a, p_b \in \mathbb{H}$ we write

$$p := p_a + \varepsilon p_b = p_0 + p_1 \mathbf{i} + p_2 \mathbf{j} + p_3 \mathbf{k} + \varepsilon(p_4 + p_5 \mathbf{i} + p_6 \mathbf{j} + p_7 \mathbf{k}) \tag{1}$$

where $\varepsilon$ is the dual unit which squares to zero, i.e. $\varepsilon^2 = 0$. Furthermore, we define the dual quaternion conjugation by $p^* := p_a^* + \varepsilon p_b^*$ and the $\varepsilon$-conjugation by $p_\varepsilon := p_a - \varepsilon p_b$. More details on the structure and multiplication rules can be found in Bottema and Roth [17, Chap. 13]. The review of the applications of dual quaternions are summarized in [18].

Given a dual quaternion $h \in \mathbb{DH}$ with $hh^* \in \mathbb{R} \setminus \{0\}$ one can prove that the map

$$\varrho_{[h]} : \mathbb{R}^3 \to \mathbb{R}^3, \quad 1 + \varepsilon x \mapsto \frac{h_\varepsilon (1 + \varepsilon x) h^*}{hh^*} \tag{2}$$

is a rigid body displacement where a point $(x_1, x_2, x_3)^T$ in $\mathbb{R}^3$ is embedded in the dual quaternions as $1 + \varepsilon(x_1 \mathbf{i} + x_2 \mathbf{j} + x_3 \mathbf{k})$. $\mathbb{DH}^\times := \{h \in \mathbb{DH} \mid hh^* \in \mathbb{R} \setminus \{0\}\}$ is called the set of *Study quaternions*. By $[h]$ we denote the set $\{\lambda h \mid \lambda \in \mathbb{R}\}$ and the map $[h] \mapsto \varrho_{[h]}$ is an isomorphism between the dual quaternions and the group of rigid body transformations SE(3). Therefore, we can identify a rigid body displacement with a point in 7-dimensional projective space $\mathbb{PR}^7$ which can be expressed as an equivalence class of 8-tuples $[\mathbf{p}] = \{\lambda(p_0, p_1, \ldots, p_7)^T \mid \lambda \in \mathbb{R}\}$ in $\mathbb{R}^8$. Because of this isomorphism we will use the term rigid body transformation and dual quaternion (or Study quaternion) interchangeably from now on. Furthermore, we will often speak of a vector or a point in the dual quaternions when we mean a representative in $\mathbb{R}^8$. Often we will refer to the canonical representative whose first entry is one (normalized by its first coordinate).

In addition to points and rigid body displacements, oriented straight lines in three space can also be represented by dual quaternions using Plücker coordinates [19, Chap. 2]. Given a direction vector $p = p_1 \mathbf{i} + p_2 \mathbf{j} + p_3 \mathbf{k}$ and a point $q = q_1 \mathbf{i} + q_2 \mathbf{j} + q_3 \mathbf{k}$ the dual quaternion $\ell = p - \frac{1}{2}\varepsilon(pq - qp)$ is called the Plücker coordinate vector. If $\ell \ell^* = 1$, then one speaks of normalized Plücker coordinates.

## 2.2. Rational motions

A one-parametric motion can be expressed as dual quaternion polynomial $C \in \mathbb{DH}[t]$. Similarly as for rigid body transformations, it is a proper motion in SE(3) if and only if $CC^* \in \mathbb{R}[t] \setminus \{0\}$. Such a motion can act on objects (points, lines, other dual quaternions, etc.), i.e. they are parameterized by the motion and follow *rational* trajectories (their position is given directly by the parameter $t$). As mentioned in the introduction, given a set of points or poses, it is possible to geometrically construct rational motions [1,2,8] and synthesize a *rational* single-loop mechanism whose points follow rational paths.

To avoid confusion for the reader, it is necessary to note that the term "motion" is also used in this study when describing trajectory (motion) planning or generation of the driving joint, i.e. when talking about joint position (rotation angle), velocity, and acceleration profiles in terms of control.

## 3. Methods

Given a 1-DoF rational mechanism characterized by a motion polynomial $C \in \mathbb{DH}[t]$ with tool (end-effector) coordinates defined as $\mathbf{p}_h \in \mathbb{DH}$ in the global frame in its "home" configuration, the following methods can be applied.

### 3.1. Direct kinematics

If the driving joint angle $\theta$ is known and the pose of the tool frame $\mathbf{p}$ needs to be determined, a function $f$ is required such that

$$\mathbf{p} = f(\theta). \tag{3}$$

The parameter $t$ parameterizes the motion $C$ in the interval $(-\infty, \infty)$ and corresponds to some angle $\theta \in [0, 2\pi)$ of the driving joint. For clarity, we derive the parameterization in terms of $t$. A rotation around an axis with unit direction vector $\vec{r} = [r_1, r_2, r_3] \in \mathbb{R}^3$ by $\theta$ can be written (see [20, Appendix B.3]) as

$$\cos\left(\frac{\theta}{2}\right) + \sin\left(\frac{\theta}{2}\right)(r_1\mathbf{i} + r_2\mathbf{j} + r_3\mathbf{k}). \tag{4}$$

We want to reparameterize in terms of $t$, i.e. achieve the same rotation by $t - q$ where $q = q_0 + q_1\mathbf{i} + q_2\mathbf{j} + q_3\mathbf{k} \in \mathbb{H}$, representing the orientation of a driving joint axis. After normalization

$$\frac{t - q}{(t - q)(t - q)^*} = \frac{t - q_0}{(t - q)(t - q)^*} - \frac{q_1\mathbf{i} + q_2\mathbf{j} + q_3\mathbf{k}}{(t - q)(t - q)^*} \tag{5}$$

we compare the *norm* of the scalar and vectorial parts to the ones of the rotation in (4) yielding

$$\cos\left(\frac{\theta}{2}\right) = \frac{t - q_0}{(t - q)(t - q)^*}, \quad \sin\left(\frac{\theta}{2}\right) = \frac{\sqrt{q_1^2 + q_2^2 + q_3^2}}{(t - q)(t - q)^*}. \tag{6}$$

From this it follows that

$$\tan\left(\frac{\theta}{2}\right) = \frac{\sqrt{q_1^2 + q_2^2 + q_3^2}}{t - q_0}. \tag{7}$$

Furthermore, one observes that the rotational axis is determined by the vectorial part of $q$, i.e. if the driving joint is given by the unit direction vector $\vec{r}$, then $[q_1, q_2, q_3] = \vec{r}$ and $q_0$ corresponds to an initial offset. Solving for $t$, we get

$$t = \frac{\sqrt{q_1^2 + q_2^2 + q_3^2}}{\tan(\frac{\theta}{2})} + q_0 \tag{8}$$

and in the inverse way, $t$ can be mapped to $\theta$ using

$$\theta = 2\arctan\frac{\sqrt{q_1^2 + q_2^2 + q_3^2}}{t - q_0}. \tag{9}$$

In the implementation, the division by zero has to be handled carefully, since $\theta = 0$ [rad] yields $t = \infty$, and $t - q_0 = 0$ yields $\theta = \pi$ [rad]. Using these equations, the *direct kinematics* problem, i.e. obtaining the desired pose of the tool frame $\mathbf{p}$, is solved with the simple equation

$$\mathbf{p} = C(t)\mathbf{p}_h \tag{10}$$

where $t$ can be replaced by (8). $C$ is the motion polynomial characterizing a mechanism and $\mathbf{p}_h$ is the displacement of the tool frame from the origin. Later in Section 4 with provided example of a Bennett mechanism, $C$ is given in (27) and $\mathbf{p}_h$ is chosen to be the identity or slightly shifted as shown in (28).

## 3.2. Inverse kinematics

If on the other hand the desired pose of the tool frame $\mathbf{p}$ is given and the angle $\theta$ of the driving joints is needed, the inverse of the function $f$ in Eq. (3) should be determined, i.e.

$$\theta = f^{-1}(\mathbf{p}). \tag{11}$$

In case of parallel robots, the inverse kinematics is mostly solved analytically using a geometric model of the robot. In case of single-loop linkages, especially those designed by factorization methods, the placement of the axes is often arbitrary, and the analytical construction of the system of equations is therefore difficult if not impossible. If a motion curve of a single-loop mechanism is known, the inverse kinematics can be solved in the dual quaternion space as an over-constrained system of eight polynomial equations with a single unknown $t$. However, if there is a numerical error in the given pose $\mathbf{p}$ (the pose on the motion curve) or the pose $\mathbf{p}$ is out of the curve (cannot be physically achieved by the linkage), the algebraic computation of such a system may fail. Therefore, we want to numerically find an approximate solution to the system of equations. As it turns out this is very similar to the standard approach for obtaining the inverse kinematics of a serial chain that uses the differential kinematics and the Newton–Raphson method [21], which can be written in the basic form as

$$\theta_{k+1} = \theta_k + \lambda \mathbf{J}(\theta_k)^\dagger \mathbf{v}_k \tag{12}$$

where $k$ is the step of iteration, $\lambda$ is the step size parameter, $\mathbf{J}^\dagger$ is the pseudoinverse of the manipulator's Jacobian, and $\mathbf{v}_k$ is an *error* to the desired pose, specified as a tool twist (generalizing the linear and angular velocity directions to the desired pose). Although this general approach would be applicable to single-loop linkages, determining the vector $\theta$ and Jacobian even for passive joints of the linkages is inefficient, especially if their parameterization by $t$ is known. Therefore, we propose a modification to find the *numerical inverse kinematics* solution tailored to the discussed linkages.

After a normalization because of the homogeneous coordinates, we can view the error $\mathbf{E}(\mathbf{t}) := \mathbf{p} - C(t) \simeq \mathbf{v}$ as a function from $\mathbb{R}$ to $\mathbb{R}^8$ by taking the coordinate vector $(p_0, p_1, \ldots, p_7)^T$ of a dual quaternion. A standard way of minimizing the error is, in terms of least squares, finding

$$\min_{t \in \mathbb{R}} \|\mathbf{E}(t)\|^2. \tag{13}$$

We may use a version of the Gauss–Newton method to find such a minimum numerically via

$$t_{k+1} = t_k + \lambda \mathbf{J}(t_k)^\dagger \mathbf{E}(t_k), \tag{14}$$

where

$$\mathbf{J}(t_k)^\dagger = \left(\mathbf{J}(t_k)^T \mathbf{J}(t_k)\right)^{-1} \mathbf{J}(t_k)^T$$
$$= \left(\sum_{k=0}^{7}\left(\frac{dC_i}{dt}(t_k)\right)^2\right)^{-1}\left(\frac{dC_0}{dt}(t_k), \ldots, \frac{dC_7}{dt}(t_k)\right) \tag{15}$$

is the pseudoinverse of the Jacobian, $t_k$ is the motion equation parameter in the $k$th step and $\mathbf{E}$ is the twist between the poses in the dual quaternion space. Compared to standard methods when using the vector of joint variables $\theta$, we easily find the pseudoinverse of the Jacobian since all joint angles of 1-DoF rational linkages depend on one single parameter $t$. Each step only requires calculating the derivative of $C$, the inner product of the derivative, and the error. Note that the derivative of $C$ can also easily be derived algebraically as it is a dual quaternion polynomial. The simplified version of the inverse kinematics problem may be written as

$$t_{k+1} = t_k + \lambda_k \frac{\dot{C}(t_k) \cdot (\mathbf{p} - C(t_k))}{\sum_{i=0}^{7} \dot{C}(t_k)_i^2} \tag{16}$$

where $\lambda_k = 1$ if $\|\mathbf{E}(t_k)\|^2 < \|\mathbf{E}(t_{k+1})\|^2$ or $\lambda_k = \lambda_{k-1}/2$.

The algorithm converges slowly when $t$ is equal to or close to $\infty$. Therefore, it is suggested to employ the following reparameterization "trick". First, search for a solution in $[-1, 1]$ and, in the event of an unsuccessful search, reparameterize $C$ by $t' = 1/t$. This maps the interval $[-\infty, -1] \cup [1, \infty]$ to the interval $[-1, 1]$. For example, solutions for valid poses at infinity will then be found at $t' = 0$.

*Example 1*

We construct a 6R linkage defining its first branch using the revolute axes $h_i$ for $i = 1, \ldots, 3$ represented as dual quaternions

$$h_1 = \mathbf{i}, \quad h_2 = 3\mathbf{k} + \epsilon\mathbf{k}, \quad h_3 = \mathbf{j} + \mathbf{k} - 2\epsilon\mathbf{k} \tag{17}$$

or, for simplification as vectorial dual quaternions in $\mathbb{PR}^7$

$$\mathbf{h}_1 = \begin{bmatrix} 0 \\ 1 \\ 0 \\ 0 \\ 0 \\ 0 \\ 0 \\ 0 \end{bmatrix}, \quad \mathbf{h}_2 = \begin{bmatrix} 0 \\ 0 \\ 0 \\ 3 \\ 0 \\ 0 \\ 0 \\ 1 \end{bmatrix}, \quad \mathbf{h}_3 = \begin{bmatrix} 0 \\ 0 \\ 1 \\ 1 \\ 0 \\ 0 \\ 0 \\ -2 \end{bmatrix}. \tag{18}$$

One can also see the Plücker vectors if one omits the first and fourth entry. The cubic rational motion $C$ is obtained by calculating

$$C(t) = (t - h_1)(t - h_2)(t - h_3) \tag{19}$$

which can be expanded and, for easier readability, expressed in the vector form as

$$\begin{bmatrix} t^3 - 4t \\ 3 - 2t^2 \\ -4t^2 - 3 \\ t \\ -7 \\ -7t \\ 2t \\ t^2 - 1 \end{bmatrix}. \tag{20}$$

We can obtain another branch via the rational motion factorization method [2]. Because of the numerical implementation we visualize results that are rounded to three decimal places

$$\mathbf{k}_1 = \begin{bmatrix} 0 \\ 1.8 \\ 2.4 \\ 0 \\ 0 \\ 0 \\ 0 \\ 0.8 \end{bmatrix}, \quad \mathbf{k}_2 = \begin{bmatrix} 0 \\ -0.723 \\ 1.215 \\ 0 \\ 0 \\ 0 \\ 0 \\ -0.492 \end{bmatrix}, \quad \mathbf{k}_3 = \begin{bmatrix} 0 \\ 0.923 \\ 0.385 \\ 0 \\ 0 \\ 0 \\ 0 \\ -1.308 \end{bmatrix}, \tag{21}$$

and due to the fact that $C(t) = (t - h_1)(t - h_2)(t - h_3) = (t - k_1)(t - k_2)(t - k_3)$ we can combine these two branches which results in a rational over-constrained 6R mechanism with 1 DoF (parameterized by $t$). For more details about this construction see [2,22].

Supposing that the driving axis is $h_1$, we choose some angle $\theta = \frac{1}{3}\pi$. It is mapped via (8) to $t = 1.732$ and the substitution in (20), i.e. solving the direct kinematics problem, yields the rigid body transformation

$$\mathbf{p} = \begin{bmatrix} -1.732 \\ -3 \\ -15 \\ 1.732 \\ -7 \\ -12.124 \\ 3.464 \\ 2 \end{bmatrix}. \tag{22}$$

This pose can be plugged in (16) as an input for solving the inverse kinematics problem (via the Gauss–Newton search), and it successfully converges in the value of $\theta = \frac{\pi}{3}$, as anticipated. Fig. 1 shows the mathematical line model of the mechanism in the pose $\mathbf{p}$. For the implementation details, see Section 4.2.

### 3.3. Trajectory planning

Joint-space motion planning and related time scaling for a smooth trajectory of a driving joint are straightforward in the case of rational linkages and standard methods such as cubic and quintic polynomial profiles can be used, as described in [20, Chap. 9]. However, sometimes a *smooth end-effector motion* is required. There, the parameterization of $C$ cannot be applied, since it is not linear, but specified by a tangent function that is also affected by $\mathbf{q}$, as previously shown in (8).

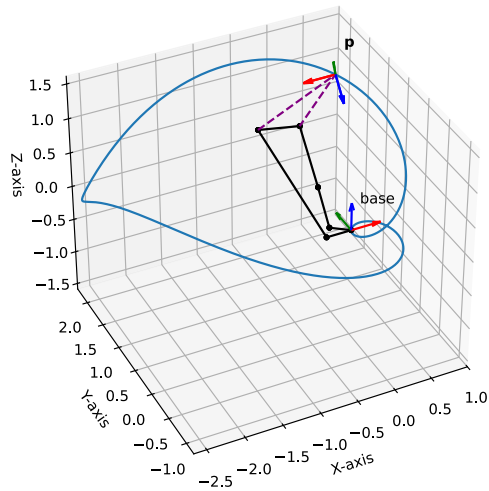

**Fig. 1.** Line-model of 6R mechanism in pose **p**; tool link is purple, revolute joints are marked by black points (points where links connect to virtual joint axes), motion path is the blue curve. The visualization was performed using Rational Linkages package [10].

The path (curve traced by the origin) of the tool frame can be expressed as a point $\mathbf{p}_t = (1, 0, 0, 0, 0, x_1, x_2, x_3)^T \in \mathbb{DH}$ and can be acted upon with a motion using (2) to produce the polynomial with zero rotation

$$\mathbf{p}_t(t) = \begin{bmatrix} x_0(t) \\ 0 \\ 0 \\ 0 \\ 0 \\ x_1(t) \\ x_2(t) \\ x_3(t) \end{bmatrix}. \tag{23}$$

The desired trajectory shall be performed between two joint coordinates $\theta_0$ and $\theta_1$, which can be mapped via (8) to $t_0$ and $t_1$. Using those values as boundaries of an integral, it is possible to determine the arc length $s$ in Euclidean 3-space $E^3$ using the standard formula

$$s = \int_{t_0}^{t_1} \sqrt{\left(\frac{d\tilde{x}_1(\tau)}{dt}\right)^2 + \left(\frac{d\tilde{x}_2(\tau)}{dt}\right)^2 + \left(\frac{d\tilde{x}_3(\tau)}{dt}\right)^2}\, d\tau \tag{24}$$

where $\tilde{x}_i(t) = x_i(t)/x_0(t)$ corresponds to the *path* of the tool frame from (23). Smooth travel of the tool can be achieved by dividing this arc into segments with equal length, i.e. equidistant travel. This approach ensures that the driving joint velocity will be adjusted to compensate for nonlinear parameterization of $C$. For the desired number of segments $n$, the length of a single segment is defined as $s_l = s(t)/n$. Starting at $t_i$ with $i = 0 \ldots n$, the next $t_{i+1}$ value is found iteratively using the bisection method with terminating condition that $s(t_{i+1}) - s(t_i) = s_l$ under a chosen threshold. Now, when the parameter values $t_{0 \ldots n}$ are obtained, the trajectory may be planned using the corresponding $\theta_{0 \ldots n}$. The driving motor velocity will be adjusted so that the travel between the segments takes approximately the same amount of time, which results, if $n$ is high enough, in a smooth motion of the tool frame.

Note that the choice of $\mathbf{p}_t$, i.e. the point that shall perform the smooth (equidistant) motion, is not limited to the tool position and can be any other point of the mechanisms, or even its virtual point in an arbitrary place. This is demonstrated in the following section. The only limit is the maximum velocity of a motor that will execute the trajectory.

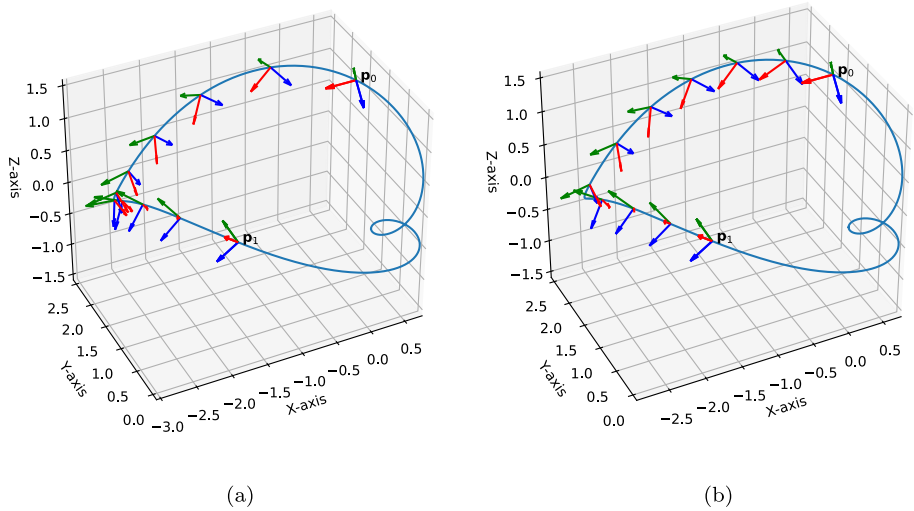

(a)  (b)

**Fig. 2.** Comparison of trajectory discretization between poses $p_0$ to $p_1$ with 10 segments: (a) linear scaling; (b) equidistant scaling.

**Table 1**
Comparison of joint angles before (linear) and after reparameterization (equidist).

| [rad] | $\theta_0$ | $\theta_1$ | $\theta_2$ | $\theta_3$ | $\theta_4$ | $\theta_5$ | $\theta_6$ | $\theta_7$ | $\theta_8$ | $\theta_9$ | $\theta_{10}$ |
|---|---|---|---|---|---|---|---|---|---|---|---|
| Linear | 1.05 | 1.41 | 1.78 | 2.15 | 2.51 | 2.88 | 3.25 | 3.61 | 3.98 | 4.35 | 4.71 |
| Equidist | 1.05 | 1.24 | 1.45 | 1.68 | 1.93 | 2.25 | 2.76 | 3.80 | 4.19 | 4.47 | 4.71 |

*Example 2*

Using the same 6R linkage as in Example 1, we choose two angles $\theta_0 = \frac{1}{3}\pi$ [rad] and $\theta_1 = \frac{3}{2}\pi$ [rad] that yield poses

$$\mathbf{p}_0 = \begin{bmatrix} -1.732 \\ -3 \\ -15 \\ 1.732 \\ -7 \\ -12.124 \\ 3.464 \\ 2 \end{bmatrix}, \quad \mathbf{p}_1 = \begin{bmatrix} 3 \\ 1 \\ -7 \\ -1 \\ -7 \\ 7 \\ -2 \\ 0 \end{bmatrix}. \tag{25}$$

The corresponding parameter values $t_0 = 1.732$ and $t_1 = -1$ are then the integral boundaries in (24), the path of the tool frame $\tilde{x}_i(t)$ is obtained from the direct kinematics (3). We decide to split the curve in $n = 10$ number of segments. Table 1 shows in the first row the linear distribution of $\theta_{0..10}$ and below the angles after applied integration and bisection, i.e. reparameterization for equidistant travel of the tool frame. In Fig. 2 the tool poses of the mechanism at the values of $\theta_{0..10}$ are visualized. In the original parameterization, the distance between poses becomes more dense around the cusp of the curve. The arc-length reparameterization divides the path in segments with approximately equal lengths (equidistant), which will provide smoother motion with respect to the tool frame. Fig. 3 then shows the position and velocity profiles of these two distributions. The velocity of the $\theta$-linear distribution is, as expected, constant. However, equidistant travel requires an increase/decrease of the driving joint velocity while going through the cusp to provide approximately constant velocity (smooth travel) of the tool frame. Note that such a peak might be challenging for a motor in real scenario and probably not advisable — this example was used to demonstrate the feasibility of the proposed methodology. In addition, the number of steps (segments) in practical application should be increased according to the specifications of a chosen motor driver, as in the next section.

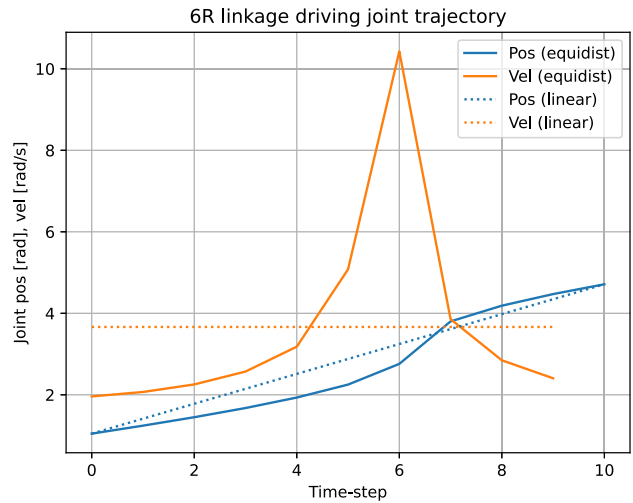

**Fig. 3.** Driving joint trajectory profiles of 6R linkage; linear and equidistant distribution for the number $n = 10$ segments.

**Table 2**
DH parameters of the Bennett linkage.

| $i$ | $d_i$ [mm] | $a_i$ [mm] | $\alpha_i$ [deg] |
| --- | --- | --- | --- |
| 0 | 0 | 192.31 | 143.14 |
| 1 | 0 | 320.57 | 90.40 |
| 2 | 0 | 192.31 | 143.14 |
| 3 | 0 | 320.57 | 90.40 |

## 4. Results

This section demonstrates the application of proposed method on Bennett mechanism synthesized from poses

$$\mathbf{p}_0 = \begin{bmatrix} 1 \\ 0 \\ 0 \\ 0 \\ 0 \\ 0 \\ 0 \\ 0 \end{bmatrix} \quad \mathbf{p}_1 = \begin{bmatrix} 1 \\ -0.208 \\ -0.033 \\ -0.069 \\ -0.006 \\ -0.014 \\ -0.045 \\ -0.026 \end{bmatrix} \quad \mathbf{p}_2 = \begin{bmatrix} 1 \\ 0.233 \\ -0.043 \\ 0.078 \\ -0.008 \\ 0.030 \\ 0.030 \\ 0.035 \end{bmatrix} \qquad (26)$$

Using the method from [1] a univariate motion polynomial $C$ is obtained in the following form

$$C(t) = \begin{bmatrix} t^2 - 0.923t + 0.226 \\ -0.116t + 0.053 \\ -0.001t - 0.010 \\ -0.039t + 0.018 \\ -0.002 \\ -0.011t + 0.006 \\ -0.020t + 0.007 \\ -0.016t + 0.008 \end{bmatrix} \qquad (27)$$

and the factorization [9] of the motion $C$ yields linkages with the Denavit–Hartenberg (DH) parameters specified in Table 2.

The visualization of poses is shown in Fig. 4. Please note that the values in Table 2 and Eqs. (26) and (27) are rounded because of the page limits and their direct usage may lead to computational issues. The exact construction of poses and precise results can be found in supplementary material (Section 4.2). For simplification, the origin frame (identity) of the mechanism is identical to the tool frame in the "home" configuration of the mechanism, that is $\theta = 0$ when $t = \infty$. Based on this input, a mechanism was constructed and its computer-aided design (CAD) model is visualized in Fig. 5.

Starting with inverse kinematics of poses $\mathbf{p}_1, \mathbf{p}_2$, the corresponding joint angles are $\theta_1 = 0.331$ [rad] and $\theta_2 = 5.893$ [rad]. Forward kinematics yields for these input angles the poses, as expected. In the next step, the trajectory planning will be performed between the two poses $\mathbf{p}_1, \mathbf{p}_2$ along the "longer" path between each other. The motion time is chosen to be 4 s and required motor control system frequency is 20 Hz, which yields 80 discrete time-steps.

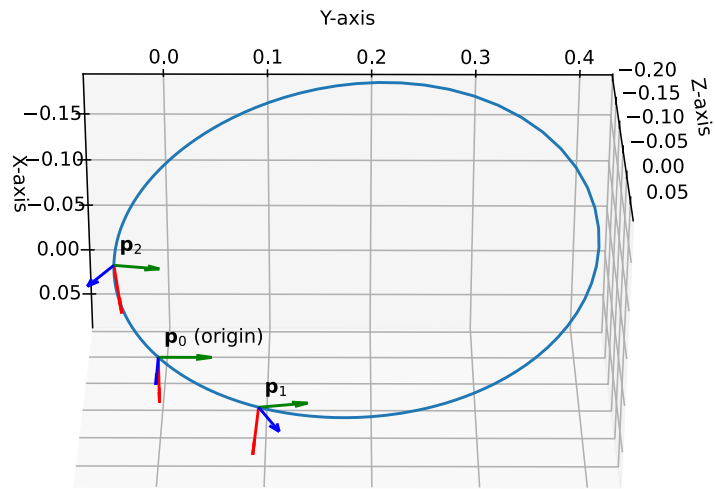

**Fig. 4.** Visualization of poses $\mathbf{p}_1$, $\mathbf{p}_2$, and origin. Motion path is the blue curve.

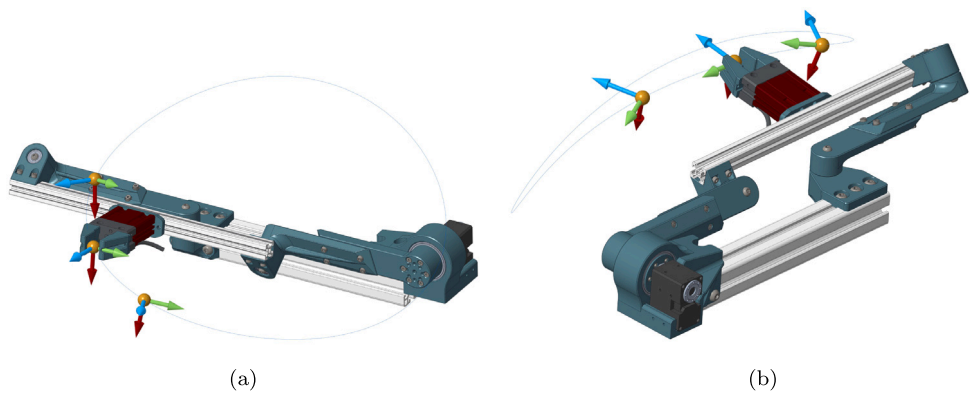

**Fig. 5.** Front (a) and back (b) views of the CAD model of Bennett mechanism in its home configuration. Poses $\mathbf{p}_1$, $\mathbf{p}_2$, and the origin $\mathbf{p}_0$ with its path are visualized.

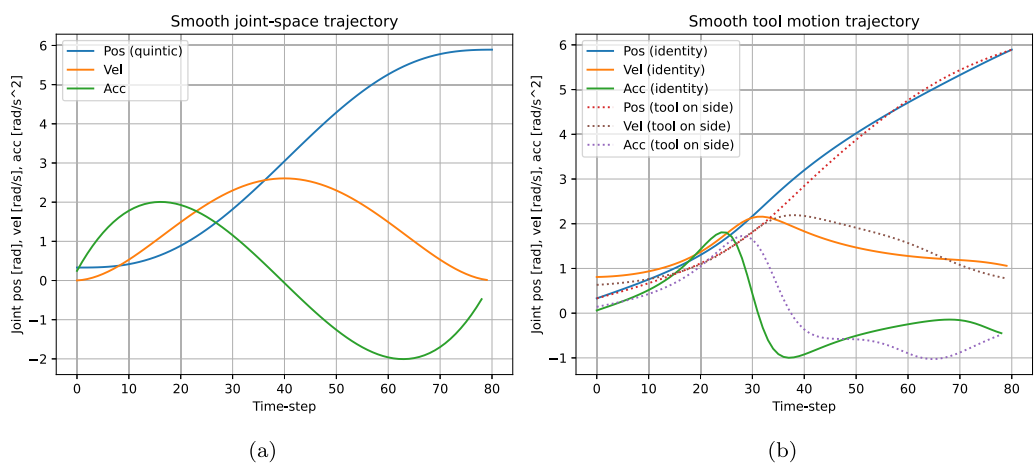

**Fig. 6.** Driving joint trajectory profiles: (a) quintic polynomial scaling; (b) equidistant path planning of the tool, in default position (identity) and in shifted position.

The straight-line trajectory planning in the joint-space has the profile shown in Fig. 6(a). For the equidistant travel, two scenarios are presented. The first profile was generated for the identity frame $\mathbf{p}_{h(\text{id})}$ and the second profile was generated for the equidistant travel of end-effector with tool-center point (TCP) placed at the side of its link, that is $-170$ mm in the direction of the $Y$-axis of origin, given as $\mathbf{p}_{h(\text{shift})}$ below.

$$\mathbf{p}_{h(\text{id})} = \begin{bmatrix} 1 \\ 0 \\ 0 \\ 0 \\ 0 \\ 0 \\ 0 \\ 0 \end{bmatrix} \quad \mathbf{p}_{h(\text{shift})} = \begin{bmatrix} 1 \\ 0 \\ 0 \\ 0 \\ 0 \\ 0 \\ 0.085 \\ 0 \end{bmatrix} \tag{28}$$

Note that you need to scale by $-\frac{1}{2}$ when embedding Cartesian points into dual quaternions, therefore $\frac{-170}{-2} = 85$ mm $= 0.085$ m. See [17, Chap. 13] for more details. From Fig. 6(b), it is possible to compare the trajectory. One can also observe that the velocity starts and ends at non-zero values, which is due to arc-length parameterization that defines the changes in motor position. This "velocity jump" may be handled by combining the two presented motion planning approaches, introducing smooth starting and braking phases for equidistant travel. Eventually, it is possible to perform a continuous full-cycle motion by planning the trajectory from 0 to $2\pi$, with the appropriate start/end velocities and rescaling.

### 4.1. Experimental setup

The proposed methodology was tested in a laboratory setup with the linkage shown in Fig. 5. The mechanism consists of an active joint driven by motor Dynamixel XH540-V270-R and 3 passive joints. All 4 joints are reinforced with ball bearings for increased rigidity. The links are made of either aluminum profiles or 3D-printed parts. The end-effector that is mounted on the mechanism is Afag EU-20 with custom grippers.

The trajectory between the two poses was performed with the following settings:

- no input to trajectory regulation (trajectory planning was done by control unit of the motor using default trapezoidal velocity profile; only start and end position angles were given as input),
- position and velocity profiles for joint-space smooth motion (interpolation by quintic polynomial) as in Fig. 6(a),
- position and velocity profiles for smooth tool motion of the end-effector in its default mounting position (TCP at origin) as in Fig. 6(a) – solid lines,
- position and velocity profiles for smooth tool motion of the end-effector shifted to side of its link, as in Fig. 6(a) – dashed lines.

The comparison can be seen in the video in the supplementary material, uploaded on [15].

### 4.2. Implementation

To simplify reproducibility, the algorithms presented were implemented in the Rational Linkages [10] Python package (version 1.8 or higher) that can be used for the synthesis and prototyping of single-loop mechanisms. The open source code is available (see [10]) on the Gitlab instance of the University of Innsbruck and all necessary information, including installation and tutorials, can be found in its documentation. The dedicated supplementary material [15] contains a Jupyter Notebook file that performs the calculations presented in this section. It can be run online through the MyBinder.org service; see the weblinks on [15].

## 5. Conclusion

Mechanisms with 1 DoF are advantageous for their simplicity in design and control, making them cost-effective and reliable for specific tasks [4]. Their predictable motion paths ensure high precision and efficiency in applications such as robotics, machinery, and automation. This study extends the current state-of-the-art of rational 1-DoF single-loop linkages in the direction of velocity motion planning in tool- and joint-space. As a key result, the numerical inverse kinematics using a variant of the Gauss–Newton method is proposed, utilizing the benefits of rationally representing the motion. The rational representation of motion simplifies the algorithm and eliminates the need of calculating the Jacobian or performing matrix operations as in the standard methods. Additionally, this paper presents a method for driving joint trajectory planning that achieves smooth motion of, preferably but not limited to, the tool. The usage of the algorithms presented was demonstrated using the Bennett mechanism; however, the methods can be directly applied to any rational single-loop linkage with 1 DoF, e.g. 6R mechanisms [2]. The results can be applied in follow-up research as input for multibody modeling [23] and the development of optimal control algorithms for this family of mechanisms.

In the future, the proposed methodology can be applied to higher degree-of-freedom linkages, which are so far rare and challenging to synthesize. However, a first example can be seen in [22]. The benefits will still apply, i.e. easy differentiation of the motion polynomial and a smaller dimension Jacobian that is due to the parameterization of the motion. One could therefore adapt the Gauss–Newton method in a straightforward way to these multi degree-of-freedom linkages.

**CRediT authorship contribution statement**

**Daniel Huczala:** Writing – review & editing, Writing – original draft, Visualization, Validation, Software, Resources, Project administration, Methodology, Investigation, Funding acquisition, Formal analysis, Conceptualization. **Andreas Mair:** Writing – review & editing, Writing – original draft, Methodology, Investigation, Formal analysis. **Tomas Postulka:** Writing – review & editing, Visualization, Validation, Software.

**Declaration of competing interest**

The authors declare the following financial interests/personal relationships which may be considered as potential competing interests: Daniel Huczala reports financial support was provided by European Commission. Tomas Postulka reports financial support was provided by Government of the Czech Republic. If there are other authors, they declare that they have no known competing financial interests or personal relationships that could have appeared to influence the work reported in this paper.

**Acknowledgments**

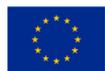

Funded by the European Union. Views and opinions expressed are however those of the author(s) only and do not necessarily reflect those of the European Union or the European Research Executive Agency (REA). Neither the European Union nor the granting authority can be held responsible for them.

Tomas Postulka was supported by project REFRESH — Research Excellence For REgion Sustainability and High-tech Industries project number CZ.10.03.01/00/22_003/0000048 via the Operational Programme Just Transition, and by specific research project SP2024/082, financed by the state budget of the Czech Republic.

**Data availability**

The supplementary material was published on Zenodo.org and referenced in the paper.

**References**

[1] K. Brunnthaler, H.-P. Schröcker, M. Husty, A new method for the synthesis of Bennett mechanisms, Int. Work. Comput. Kinemat. (2005).
[2] G. Hegedüs, J. Schicho, H.-P. Schröcker, Four-pose synthesis of angle-symmetric 6R linkages, J. Mech. Robot. 7 (4) (2015) http://dx.doi.org/10.1115/1.4029186.
[3] Y. Li, G. Bone, Are parallel manipulators more energy efficient? in: Proceedings 2001 IEEE International Symposium on Computational Intelligence in Robotics and Automation (Cat. No.01EX515), in: CIRA-01, IEEE, 2001, http://dx.doi.org/10.1109/cira.2001.1013170.
[4] T. Postulka, D. Huczala, S. Weerasinghe, G. Orzechowski, A. Mikkola, Z. Bobovsky, R. Hunady, Applicability and effectiveness of the Bennett mechanism, 2025, http://dx.doi.org/10.5281/zenodo.15095819, Preprint Zenodo.
[5] Z. Li, G. Nawratil, F. Rist, M. Hensel, Invertible paradoxic loop structures for transformable design, Comput. Graph. Forum 39 (2) (2020) 261–275, http://dx.doi.org/10.1111/cgf.13928.
[6] M. Pfurner, A new family of overconstrained 6R-mechanisms, Proceedings of EUCOMES 2008, 2008, pp. 117–124, http://dx.doi.org/10.1007/978-1-4020-8915-2_15.
[7] B. Jüttler, Über zwangläufige rationale bewegungsvorgänge, Österreich. Akad. Wiss. Math.- Nat. Kl. S.- B. II 202 (1–10) (1993) 117–232.
[8] S. Zube, Interpolation method for quaternionic-Bezier curves, Liet. Mat. Rink. 59 (2018) 13–18, http://dx.doi.org/10.15388/lmr.a.2018.03.
[9] G. Hegedus, J. Schicho, H.-P. Schröcker, Factorization of rational curves in the Study quadric, Mech. Mach. Theory 69 (2013) 142–152, http://dx.doi.org/10.1016/j.mechmachtheory.2013.05.010.
[10] D. Huczala, J. Siegele, D.A. Thimm, M. Pfurner, H.-P. Schröcker, Rational linkages: From poses to 3D-printed prototypes, in: Advances in Robot Kinematics 2024. ARK 2024, Springer International Publishing, 2024, http://dx.doi.org/10.1007/978-3-031-64057-5_27.
[11] G. Hegedüs, J. Schicho, H.-P. Schröcker, Construction of overconstrained linkages by factorization of rational motions, in: Latest Advances in Robot Kinematics, Springer Netherlands, 2012, pp. 213–220, http://dx.doi.org/10.1007/978-94-007-4620-6_27.
[12] O. Cakar, A.K. Tanyildizi, Application of moving sliding mode control for a DC motor driven four-bar mechanism, Adv. Mech. Eng. 10 (3) (2018) 168781401876218, http://dx.doi.org/10.1177/1687814018762184.
[13] A. Müller, Generic mobility of rigid body mechanisms, Mech. Mach. Theory 44 (6) (2009) 1240–1255, http://dx.doi.org/10.1016/j.mechmachtheory.2008.08.002.
[14] H.-S. Yan, G.-J. Yan, Integrated control and mechanism design for the variable input-speed servo four-bar linkages, Mechatronics 19 (2) (2009) 274–285, http://dx.doi.org/10.1016/j.mechatronics.2008.07.008.
[15] D. Huczala, A. Mair, T. Postulka, DK, IK, and motion planning of rational linkages (supplementary material), 2024, http://dx.doi.org/10.5281/zenodo.14411792.
[16] J.M. Selig, Rational interpolation of rigid-body motions, in: Advances in the Theory of Control, Signals and Systems with Physical Modeling, Springer Berlin Heidelberg, 2010, pp. 213–224, http://dx.doi.org/10.1007/978-3-642-16135-3_18.
[17] O. Bottema, B. Roth, Theoretical Kinematics, North-Holland Publishing Company, 1979.
[18] J.G. Farias, E.D. De Pieri, D. Martins, A review on the applications of dual quaternions, Machines 12 (6) (2024) 402, http://dx.doi.org/10.3390/machines12060402.

[19] H. Pottmann, J. Wallner, Computational line geometry, Springer, 2001, http://dx.doi.org/10.1007/978-3-642-04018-4,
[20] K.M. Lynch, F.C. Park, Modern Robotics: Mechanics, Planning, and Control, Cambridge University Press, 2017.
[21] A. Goldenberg, B. Benhabib, R. Fenton, A complete generalized solution to the inverse kinematics of robots, IEEE J. Robot. Autom. 1 (1) (1985) 14–20, http://dx.doi.org/10.1109/jra.1985.1086995.
[22] J. Frischauf, M. Pfurner, D.F. Scharler, H.-P. Schröcker, A multi-bennett 8R mechanism obtained from factorization of bivariate motion polynomials, Mech. Mach. Theory 180 (2023) 105143, http://dx.doi.org/10.1016/j.mechmachtheory.2022.105143.
[23] J. Gerstmayr, Exudyn – a vc++-based Python package for flexible multibody systems, Multibody Syst. Dyn. (2023) http://dx.doi.org/10.1007/s11044-023-09937-1.